\documentclass[10pt,journal,compsoc]{IEEEtran}
\usepackage{cite}
\usepackage[pdftex]{graphicx}
\usepackage{algorithm}
\usepackage{algpseudocode}
\usepackage{comment}
\usepackage[normalem]{ulem}
\usepackage{caption}
\usepackage{subcaption}
\usepackage{booktabs}
\usepackage{multirow}
\usepackage[frozencache,cachedir=.]{minted}
\usepackage{booktabs}
\usepackage{xcolor}
\usepackage{url}

\begin{document}
%
\title{Parallel Blockwise Knowledge Distillation for Deep Neural Network Compression}
%
%
%
%

\author{Cody Blakeney,
        Xiaomin Li,
        Yan Yan,
        Ziliang Zong
        }
\IEEEtitleabstractindextext{%
\begin{abstract}
Deep neural networks (DNNs) have been extremely successful in solving many challenging AI tasks in natural language processing, speech recognition, and computer vision nowadays. However, DNNs are typically computation intensive, memory demanding, and power hungry, which significantly limits their usage on platforms with constrained resources. Therefore, a variety of compression techniques (e.g. quantization, pruning, and knowledge distillation) have been proposed to reduce the size and power consumption of DNNs. Blockwise knowledge distillation is one of the compression techniques that can effectively reduce the size of a highly complex DNN. However, it is not widely adopted due to its long training time. In this paper, we propose a novel parallel blockwise distillation algorithm to accelerate the distillation process of sophisticated DNNs. Our algorithm leverages local information to conduct independent blockwise distillation, utilizes depthwise separable layers as the efficient replacement block architecture, and properly addresses limiting factors (e.g. dependency, synchronization, and load balancing) that affect parallelism. The experimental results running on an AMD server with four Geforce RTX 2080Ti GPUs show that our algorithm can achieve 3x speedup plus 19\% energy savings on VGG distillation, and 3.5x speedup plus 29\% energy savings on ResNet distillation, both with negligible accuracy loss. The speedup of ResNet distillation can be further improved to 3.87 when using four RTX6000 GPUs in a distributed cluster.     


\end{abstract}

}

\maketitle

\IEEEdisplaynontitleabstractindextext

%
\IEEEpeerreviewmaketitle

\IEEEraisesectionheading{\section{Introduction}\label{sec:introduction}}

\IEEEPARstart{D}{eep} neural networks (DNNs) have successfully solved many challenging tasks in natural language processing, speech recognition, and computer vision. They have been widely used to support various exciting and powerful applications such as translating social media messages into hundreds of languages\cite{10.1145/3173574.3173791, wu2016google}, creating more interesting image previews \cite{theis2018faster},  providing virtual green screen backgrounds for video calls, and helping amateurs take professional looking photos \cite{wadhwa2018synthetic, alhashim2018high}. However, DNNs are typically computation intensive, memory demanding, and power hungry, which prevents them to be ubiquitously deployed on edge devices (e.g. mobile phones and IoT equipment). Therefore, it is vital to develop compression techniques that can significantly reduce the computation, size, and power consumption of DNNs in order to deploy them on resource constrained systems. 

Currently, the state-of-the-art compression techniques for DNNs can be broadly divided into four categories: (1) Quantization \cite{gong2014compressing, wu2016quantized, vanhoucke2011improving, han2015deep}, which reduces the memory footprint and computation demand of DNNs by representing each weight using less number of bits. (2) Low-rank approximation\cite{denton2014exploiting, jaderberg2014speeding, lebedev2014speeding}, which uses matrix/tensor decomposition to estimate the informative parameters that can be applied to both convolutional layers and fully connected  layers. (3) Network pruning, which includes unstructured pruning ~\cite{han2015learning} and structured pruning ~\cite{structured-pruning}. Both aim to eliminate unimportant weights in a DNN to reduce its size and computation demand. The difference is that unstructured pruning sets unimportant weights to zero in an arbitrary way while structured pruning can remove the entire kernel or filter when necessary. (4) Knowledge distillation ~\cite{hinton2015distilling}, which distills the knowledge learned from a large complex teacher model (or an ensemble of models) to a simple student model with less computation,  smaller size, and similar accuracy. Techniques that belong to the first two categories have been extensively studied and widely adopted. For network pruning, unstructured pruning tends to have better accuracy but it needs special hardware support to exploit weight sparsity \cite{parashar2017scnn, han2017ese, zhang2016cambricon}. Structured pruning can be deployed on existing hardware but it suffers from lower compression ratios or greater accuracy loss. Knowledge distillation can achieve good accuracy, run on both CPUs and GPUs, and are compatible with most existing compression techniques, which make it a very promising technique for DNN compression.  

Nevertheless, the training process of knowledge distillation is particularly slow because it requires running two full copies of DNNs simultaneously to let the teacher model instruct the student model. It often takes nearly as long as training from scratch \cite{hinton2015distilling}, which is the primary reason why knowledge distillation is not widely adopted in practice. Meanwhile, knowledge distillation strives to create a simplified student model that could be totally different from the original teacher model as long as accuracy is preserved. Therefore, the effectiveness of model distillation is often challenged in the aspects of teacher-student network optimization and student network structure design ~\cite{wang2018progressive}. Hinton et al. addressed this problem by defining the architecture of the  student model in advance and training the entire student model as a whole \cite{hinton2015distilling}. This further slows down the training process because the training of latter layers will highly depend on the training results of the front layers. Wang et al. improved this process by proposing the progressive blockwise knowledge distillation method, which can distill the knowledge of the entire teacher model by locally extracting the knowledge of each block in a progressive learning manner\cite{wang2018progressive}. Although it alleviates the dependency problems, their method is essentially still a sequential distillation process and takes long time to train.

\begin{figure*}
  \begin{center}
    \vspace{0.35in}
    \includegraphics[width=\textwidth]{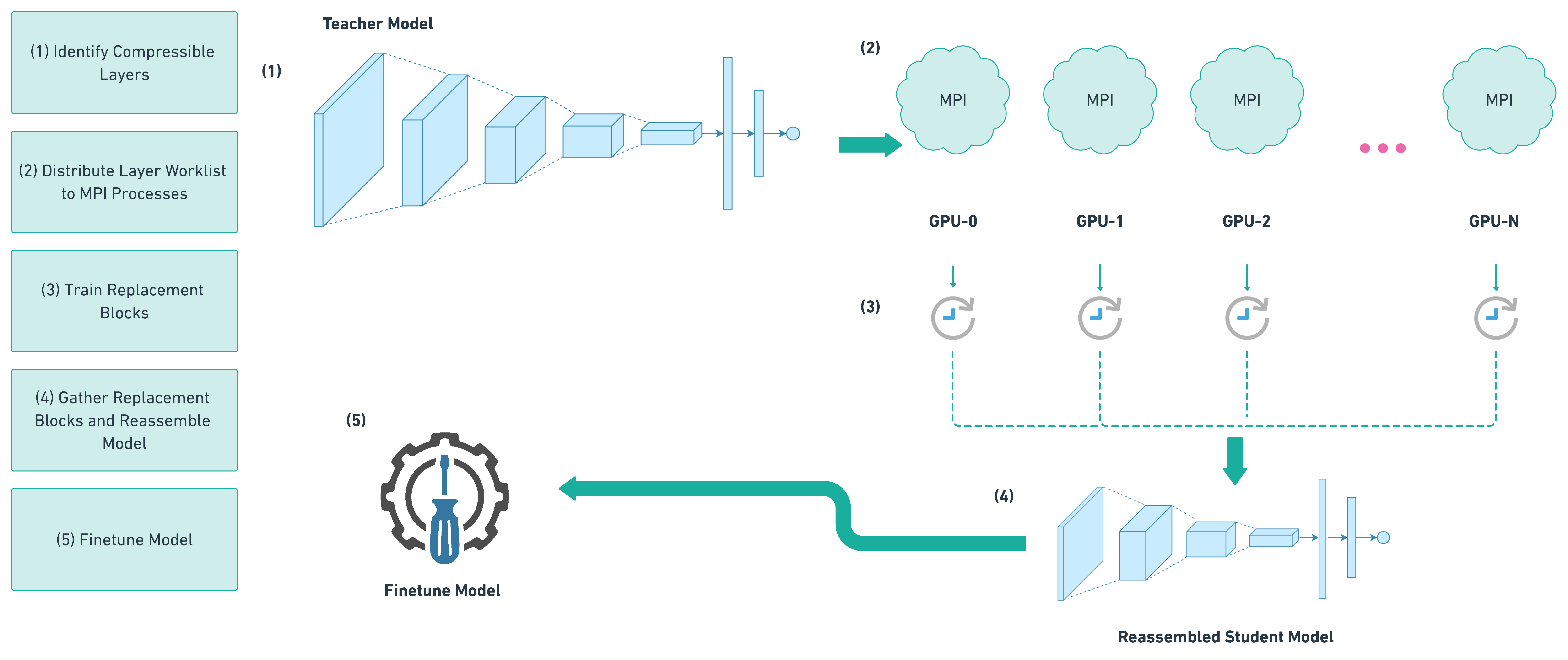}
  \caption{System level overview of parallel blockwise distillation}
  \label{fig:overview}
  \end{center}
  \vspace{-0.2in}
\end{figure*}

In this paper, we propose a novel parallel algorithm for blockwise knowledge distillation, which works as illustrated in Figure \ref{fig:overview}. First it identifies all compressible layers in the teacher model and creates independent tasks for replacing these layers (one task for each layer). All tasks will be distributed to multiple GPUs via a group of TensorFlow instances and MPI processes by following a specific scheduling algorithm such as round robin, bin packing, or work stealing. Each process trains their replacement blocks independently on the activations of the layers to be replaced. Once all layers have been trained, the main MPI process gathers the weights from all processes and reassembles the compressed DNN. Last but not the least, the reassembled student model will be fine-tuned to minimize the accuracy loss from the teacher model. 


Our algorithm has the following advantages: (1) It requires relatively few epochs for each training layer and runs once (not iteratively), which reduces training time over traditional compression methods. (2) It utilizes task parallelism to increase the simultaneous execution of different tasks on multiple GPUs with very little communication, which minimizes the overhead of synchronization. (3) It transparently ensures the elasticity and scalability because users do not need to adjust their hyper parameters when more GPUs are added. (4) It significantly reduces both training time and total energy consumption, which is crucial for DNNs used in production software where they are often deployed, evaluated, retrained, and redeployed in repeated development cycles. 

The remainder of the paper is organized as follows. Section 2 discusses related work in compression techniques for DNNs and highlights the uniqueness of our solution. Section 3 explains the proposed parallel blockwise distillation algorithm in details. Section 4 presents the system configuration and illustrates the comprehensive results of numerous experiments that we conduct using various models and datasets. Finally, Section 5 concludes this study and discusses future work.


\section{Related Work}

The great success of DNNs is undoubtedly at the cost of excessive amount of computation, memory, energy, and carbon emission. For example,  the  computation  cost  of  deep  learning  algorithms, from  AlexNet  in  2012  to  recent  AlphaGo  Zero  in  2019, has increased  by  300,000x  in  6  years \cite{DarioPost}.  Training the transformer model (213M parameters) of Natural Language Processing with neural architecture search consumes 656,347 kwh of energy and emits as much carbon as five cars in their lifetimes \cite{strubell2019energy}.  This is clearly not feasible for deploying such DNNs on edge devices and even not sustainable to train them in the cloud. Therefore, it becomes highly desirable to compress DNNs to smaller models before deploying them. Equally important, the compression process itself has to be efficient without compromising accuracy. In this section, we discuss the state-of-the-art compression techniques for DNNs and the literature to accelerate the training process of DNNs via parallelization.


\subsection{Deep Neural Network Compression}

Compression techniques for DNNs can be broadly divided into three categories, which include quantization, network pruning, and knowledge distillation. We will briefly summarize each category but it is worth noting that there are certainly other techniques that do not belong to the three categories thus are not included in our discussions.   

\subsubsection{Quantization}

Mathematically, quantization refers to the process of reducing the number of bits that represent a number. In the early stage of deep learning, the predominant numerical format used in DNNs is 32-bit floating points (FP32), which provides high precision but is computationally  expensive. If the accuracy of deep learning can be preserved, quantization provides obvious benefit of faster calculation, less energy consumption, and smaller memory footprint. This motivated extensive research in quantization-aware training and learning. Gong et al. applied k-means scalar quantization to DNNs and achieved 16-24 times compression with only 1\% loss on accuracy \cite{gong2014compressing}. Wu et al. quantized the weights of both convolutional and fully-connected layers and observed 4-6x speed-up and 15-20x compression with merely 1\% drop on accuracy \cite{wu2016quantized}. Gupta et al. used FP16 representation in CNN training and observed little to no degradation in accuracy \cite{gupta2015}. Gysel et al. verified that taking a model trained for FP32 and directly quantizing it to INT8 can result in similarly accuracy with some fine-tuning \cite{gysel2018}. 
In most cases, re-training is necessary in order to gain reasonable accuracy if ultra-low precision weights are used. 
Generally speaking, quantization is straightforward and easy to implement. Moreover, it is compatible with other compression techniques such as pruning and knowledge distillation. Therefore, it has been widely used as an effective compression technique today.   


\subsubsection{Network Pruning}

The key idea of network pruning is to reduce model size and computation time by eliminating or masking unimportant weights and activations of neural networks. Based on different granularity, pruning can be further categorized as structured pruning and unstructured pruning. Structured pruning eliminates the entire group of elements at the kernel or filter level so it is also referred to as course-grained pruning\cite{mao2017exploring}. On the other hand, unstructured pruning is more fine-grained by masking individual weight. Typically, unstructured pruning can achieve better accuracy and a higher compression ratio at the cost of inducing sparsity. It requires special hardware that can support the irregularity of the sparse computation \cite{parashar2017scnn, han2017ese, zhang2016cambricon}. Structured pruning can be deployed on existing hardware but it suffers from lower compression ratios or greater accuracy loss.   



Once a pruning ratio is determined, there are two different strategies to prune a DNN. One is to concatenate tensors of each layer as whole, prune the network from a global view to reach the target ratio. This strategy can often lead to higher accuracy but requires longer pruning time and a more complex pruning schedule. Another strategy is to prune each layer to a preset sparsity and make sure the average sparsity of all layers equals to the target pruning ratio. According to the number of iterations, a DNN can be pruned once or iteratively. One-shot pruning is fast but pruned parameters lose the opportunity for re-considering their importance. Iterative pruning is followed by retraining, which keeps tracking the importance of pruned parameters and recovers their values once are evaluated as important parameters again. Iterative pruning usually has better accuracy but it takes a longer time to prune due to the fine-tuning process. 



\subsubsection{Knowledge Distillation}
Knowledge distillation compresses a DNN by training a small student model to mimic a larger teacher model (or ensemble of models). If the training is successful, the knowledge learned from the cumbersome teacher model will be distilled to the simplified student model without compromising accuracy. This method was first invented by Bucila et al.\cite{bucilua2006model} and gained much attention after Hinton et al. generalized this idea in \cite{hinton2015distilling}. The knowledge is transferred from the larger model to the smaller model by minimizing a loss function where the target is the output of a softmax function (class probabilities) on the large model's logits. Knowledge distillation can be combined with other model compression techniques such as quantization \cite{tann2017hardware} \cite{mishra2017apprentice} \cite{polino2018model} and pruning \cite{theis2018faster} \cite{ashok2017n2n}. 

The conventional knowledge distillation method \cite{hinton2015distilling} relies on the whole teacher models to generate student models, which can be viewed as global-wise distillation. This strategy needs a huge search space of the student network with a wide variety of network configurations that are intractable and unstable in real practice. To tackle this problem, Wang et al. proposed a progressive blockwise distillation method that can distill the knowledge of the entire teacher network by locally extracting the knowledge of each block and transfer to student sub-networks \cite{wang2018progressive}. Although their method reduces the training time, the distillation process is still conducted in a serial manner. We take one step further in this paper to propose a novel parallel algorithm for blockwise knowledge distillation.

\subsection{Parallel and Distributed Deep Learning}

As the number of parameters increases dramatically, training a sophisticated DNN on a large dataset takes a substantial amount of time and energy. Apparently, deep learning has quickly emerged as a high-performance computing (HPC) problem due to its massive volume of computation and data processing. Parallelization is the key to accelerate HPC applications. However, parallelizing DNNs is a daunting task because: (1) the dataset might be too big to fit into memory; and (2) the dependencies among layers (i.e. the latter layer inputs depends on former layer outputs) significantly limit parallelization opportunities.  

Ben-Nun et al. did a comprehensive survey on parallel and distributed deep learning \cite{BenNun2019} and categorized existing algorithms as data parallelism, model parallelism, and pipelining. Data parallelism partitions the work of the batch samples among multiple cores or devices. It also helps to alleviate the problem of the dataset being too big to fit into the memory of a single device. However, batch-splitting (data-parallelism) requires users to carefully tune their hyper-parameters \cite{keskar2016large} and could result in worse accuracy \cite{masters2018revisiting}. Additionally, it does not scale well beyond single device systems because of the communication cost to average gradients calculated from all the small batches. Model parallelism divides a DNN into pieces and allocates one or multiple consecutive layers to a single device to calculate its gradients. Nevertheless, the interdependencies among layers generate high communication costs, which greatly affects the possible speedup. To mitigate this problem, Coates et al. used Locally Connected Networks (LCNs) \cite{Coates2013} and Lee et al. proposed replication-based TreeNets \cite{Lee2015}. Günther et al. explored multiple layer-parallel methods to accelerate the training process of ResNet by replacing the sequential forward and backward propagation using a parallel nonlinear multigrid \cite{gunther2020layer}. Pipelining can be viewed as a combination of data parallelism and model parallelism. It tries to overlap computation between layers and increase the utilization of multiple devices. However, pipelining requires the right data must arrive at the right time (hard to control in practice) and it does not scale well because the latency is proportional to the number of incurred devices.

In the context of knowledge distillation, the original method proposed by Hinton et al. was completely serial with high interdependencies among layers. Wang et al. improved Hinton's method by leveraging local training and progressive blockwise learning~\cite{wang2018progressive}. Although Wang's method relaxed the dependencies among layers, the distillation process was still conducted in a sequential pattern. 
Our proposed algorithm makes knowledge distillation an embarrassingly parallel process by eliminating almost all dependencies (only two synchronizations are necessary at the beginning and the end of the process). It achieves excellent speedup and scalability without compromising accuracy or requiring users to modify their hyper-parameters when more devices are added.

\section{Methodology}

\subsection{Independent Blockwise  Distillation}\label{indep_replacement}

Progressive blockwise distillation \cite{wang2018progressive} works by defining groups of layer blocks from a teacher model and creating a less computationally intense set of layers to replace them. Let us consider a teacher network $T$ that can be represented as a composite function of its $k$ subnetwork blocks:

\begin{table}[!h]
\centering
\caption{Summary of Notations}
\label{tab:terms}
\begin{tabular}{@{}ll@{}}
\toprule
Term                                  & Definition                           \\ \midrule
$T$                                   & Teacher Model                        \\
$S$                                   & Student Model                        \\
block                                 & sequential group of 1 or more layers \\
$L^{k}_{local}$                       & Local loss for block k               \\
$L^{k}_{cls}$                         & Cross Entropy Loss                   \\
$\lambda_{local}$                     & hyper-parameter used in \cite{wang2018progressive} \\
                                      &                                      \\ \bottomrule
\end{tabular}
\end{table}
\begin{equation}
    T = c \circ f_{k} \circ f_{k-1} \circ ... \circ f_{1}
\end{equation}

The goal of knowledge distillation is to derive a smaller student network $S$ where its $k$ network blocks are simplified networks but can mimic the larger networks in the teacher model.

\begin{equation}
    S = c \circ g_{k} \circ g_{k-1} \circ ... \circ g_{1}
\end{equation}

In \cite{wang2018progressive}, the distillation process is conducted progressively in a "bottom up" fashion, which refers to replacing the blocks in the order of (1, 2, ..., $k$). The blocks are trained on both the local activations of the blocks being replaced in the teacher model as well as the cross entropy loss from the student with the ground truth labels. Once a layer is trained, it is frozen in the student model and the next replacement layer is trained. This requires each layer to be trained sequentially.

Let us consider the case of replacing block $k$ in our model. Let $f_k$ be the function that maps an input image to the activations at block $k$. Let $g$ be the replacement block for block $k$. The local loss function can be represented as:

\begin{equation}\label{layer_loss}
    L^{k}_{local} = \frac{1}{N} \sum_{i = 1}^{N} \vert\vert f_k(x_i) - g_{k} \circ f_{k - 1}(x_i) \vert\vert^2
\end{equation}

and the combined loss would be defined as:

\begin{equation}
    L^k = \lambda_{local} L^{k}_{local} + L^{k}_{cls}
\end{equation}

where $\lambda_{local}$ is a hyper-parameter to balance the loss terms. $x_i$ is an individual training image and $N$ is the total number of training images.

We improved the progressive blockwise distillation method \cite{wang2018progressive} from the following perspectives. First, we consider only the local loss (equation \ref{layer_loss}) between the student model and the teacher model's feature maps. This makes it unnecessary to store a full copy of either the teacher or the student model into GPU memory. We can simply recreate the subset of teacher model layers up until and including the block that is to be replaced. The activations from the  proceeding block are used as inputs for the student block as described in equation \ref{layer_loss}, and the activations from the original block being replaced become our labels. 

Second, instead of replacing the blocks in a progressive bottom up fashion, all candidate blocks are trained independently and simultaneously. This not only eliminates most unwanted dependencies but also allows us to reduce the accumulated error from the model. Once every replacement block is trained, the algorithm evaluates each one of them to see if they reach or exceed a predetermined accuracy threshold. Only the blocks that meet the predetermined accuracy threshold will be added to the reconstructed model. Then, the final student model will be produced after some fine-tuning to regain accuracy.

\subsection{Replacement Block Architecture}

\begin{figure*}[!h]
\centering
\begin{subfigure}{.23   \textwidth}
    \centering
    \includegraphics[height=.23\paperheight]{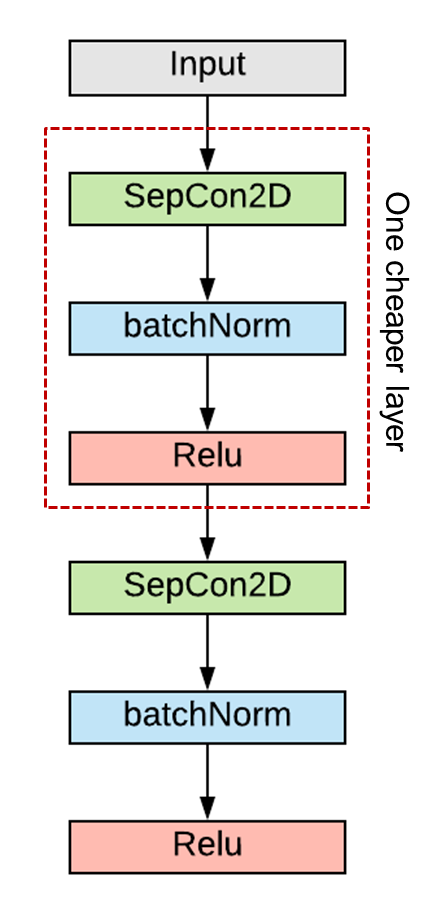}
    \caption{Two Layer Replacement}
\end{subfigure}
 \begin{subfigure}{.23\textwidth}
     \centering
     \includegraphics[height=.23\paperheight]{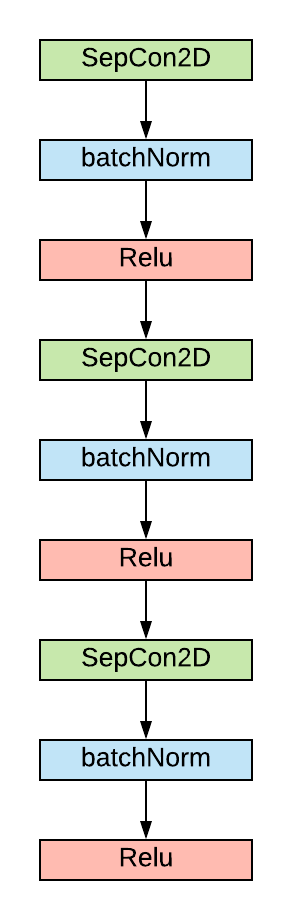}
     \caption{Three Layer Replacement}
 \end{subfigure}
\begin{subfigure}{.23\textwidth}
    \centering
    \includegraphics[height=.23\paperheight]{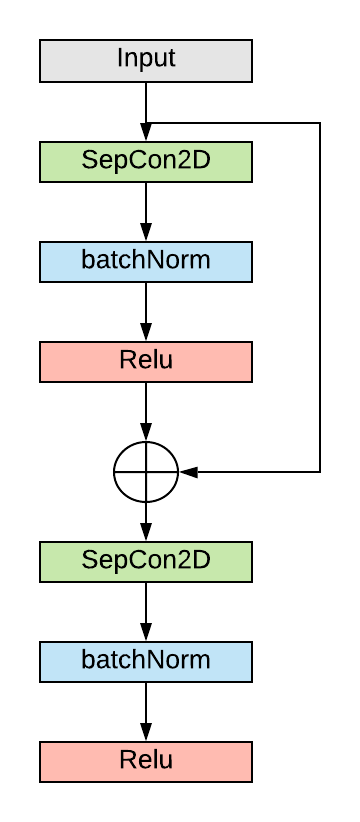}
    \caption{Two Layers w/ Skip }
\end{subfigure}%
 \begin{subfigure}{.23\textwidth}
     \centering
     \includegraphics[height=.23\paperheight]{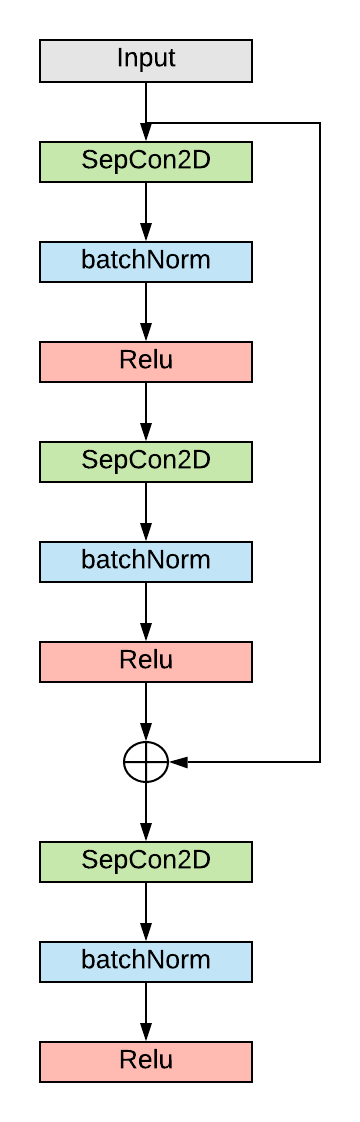}
     \caption{Three Layers w/ Skip }
 \end{subfigure}

\caption[short]{Candidate replacement block architectures. Note that we refer to the combination of depthwise separable, batch normalization, and relu activation as one layer.}
\label{fig: can}
\end{figure*}

\begin{figure*}
  \begin{center}
    \vspace{0.35in}
    \includegraphics[width=\textwidth]{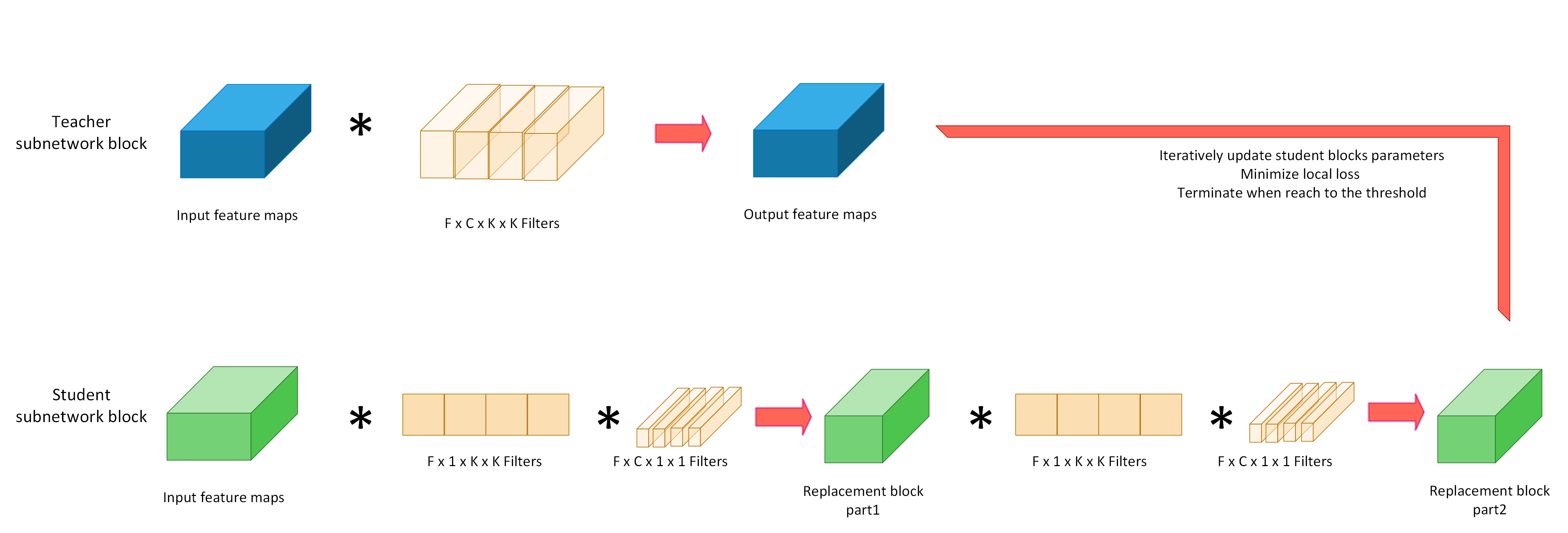}
  \caption{Expanded view of two layer replacement block with depthwise separable convolutions}
  \label{fig:block_replacement}
  \end{center}
  \vspace{-0.2in}
\end{figure*}

Finding an efficient block for the student model to replace the expensive block in the teacher model is the fundamental step of blockwise knowledge distillation. 
Most existing work on blockwise distillation \cite{wang2018progressive, gao2018embarrassingly, kulkarni2019stagewise} focus on replacement of regular repeated sections of a model. They are often dependent on the specific architecture of the teacher model thus do not generalize well to other types of models. Moreover, the larger blocksize results in less overall available parallelism. Therefore, our algorithm seeks for replacement block strategies that facilitate parallelism and generalize well to many different models without significant modification by users.

Google's prior study \cite{howard2017mobilenets} revealed that for a given convolutional layer with a 3x3 kernel size, the equivalent depthwise separable layer that produces a feature map with identical dimensions only requires 1/9 of the number of calculations and 1/9 of the amount of memory. Theoretically, computation and memory usage can be further reduced if multiple depthwise separable layers are stacked to build the replacement blocks in the student model. Inspired by their work, we explored to use depthwise separable layers as replacement block architecture in \cite{cody2020}. Since the total amount of FLOPS and parameters in an ideal replacement block must remain much smaller than traditional convolution layers, our choice is limited to two or three depthwise separable layers, which have 2/9 or 1/3 of FLOPS/parameters respectively. Specifically, we investigated four candidate architectures (see Figure~\ref{fig: can}), which include the two depthwise separable layers (a), three depthwise separable layers (b), two depthwise separable layers with skip connections similar to residual blocks (c), and three depthwise separable layers with skip connections (d). We designed experiments to evaluate the effectiveness of four candidates as follows. First, we trained a VGG16 model on CIFAR10 with the baseline accuracy of 86.45\%. Each candidate architecture aims to replace the second convolution layer in the VGG model. We trained each block for 20 epochs on the teacher model's intermediate activations. After that, the layers were inserted into the original model to replace layer 2 and the accuracy of the model was recorded. The weights of the 2nd layer were then fine-tuned using traditional cross entropy loss on the labeled CIFAR10 image data. During fine-tuning all other model weights other than the replacement block were frozen. 

Table 1 summarizes the results of the four candidate architectures, from which we can clearly see that the simple architecture with two depthwise separable layers (i.e. candidate a) yields the best accuracy. Therefore, we select this architecture as the replacement block architecture throughout the rest of the experiments. Figure ~\ref{fig:block_replacement} illustrates the expanded view of the selected replacement block architecture with two depthwise separable layers.

\begin{table}[h]
\centering
\caption{Top-1 Accuracy and Fine Tuning Top-1 Accuracy results of different replacement architectures.}
\begin{tabular}{@{}llll@{}}
\toprule
Candidate Architectures                                                             & Top-1   & FT Top-1 \\ \midrule
\multicolumn{1}{l|}{Two Layer}        & 86.40\% & 87.32\%  \\
\multicolumn{1}{l|}{Three Layer}      & 85.28\% & 85.37\%  \\
\multicolumn{1}{l|}{Two Layer w/Skip}    & 86.28\% & 86.98\%  \\
\multicolumn{1}{l|}{Three Layer w/Skip}  & 86.16\% & 87.03\%  \\ \bottomrule
\end{tabular}

\label{tab:my-table}
\end{table}

\begin{figure*}
  \begin{center}
    \vspace{0.35in}
    \includegraphics[width=\textwidth]{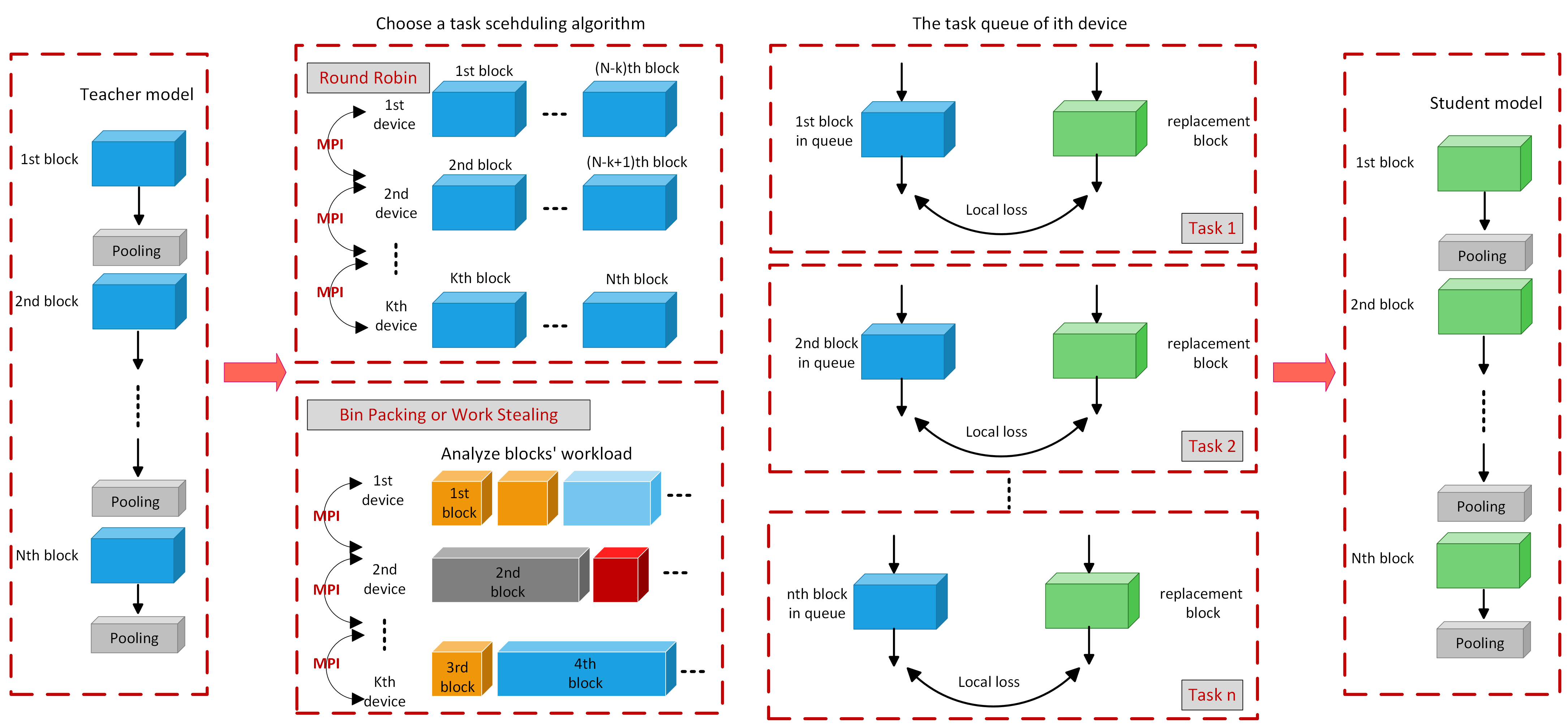}
  \caption{Overview of the parallel blockwise distillation process}
  \label{fig:parallel_replacement}
  \end{center}
  \vspace{-0.2in}
\end{figure*}

\subsection{Parallel Blockwise Distillation}
Once the blockwise distillation process and replacement block architecture are determined, the last missing puzzle is how to parallelize the training process. 

\begin{algorithm}
\caption{Parallel Blockwise Distillation}\label{blockwise}
    \hspace*{\algorithmicindent} \textbf{Input} : Teacher Network $ T = c \circ f_{k} \circ f_{k-1} \circ ... \circ f_{1}$\\
    \hspace*{\algorithmicindent} \textbf{Output} : Student Network $ S = c \circ g_{k} \circ g_{k-1} \circ ... \circ g_{1}$
\begin{algorithmic}[1]
        \State identify all layers to be compressed
        \State broadcast layer assignment to workers
        
        \For{each worker}
            \ForAll{allocated layers}
                \State train each layer as described in \ref{indep_replacement}
                \If{epoch $\%2 == 0$}
                    \State evaluate model with student block
                \EndIf
            \EndFor
        \EndFor
        
        \State  gather all weights to process 0
        
        \ForAll{layers}
            \If{accuracy $>$ threshold}
                \State replace teacher block with student block
            \EndIf
        \EndFor
        
        \State fine-tune resulting model \\
        \Return S

\end{algorithmic}
\end{algorithm}

Algorithm \ref{blockwise} explains the parallel blockwise distillation method in detail. The first step is to identify all layers of the model to be replaced. Our algorithm considers every convolutional layer with the exception of 1x1 convolutions. This is because the 1x1 convolution is already more efficient than depthwise separable convolution. As depicted in Figure \ref{fig:parallel_replacement}, each identified replacement block is viewed as an independent task and all tasks will be allocated to the available GPUs based on a selected scheduling algorithm. Users can choose either a naive scheduling algorithm (e.g. round robin) or a more advanced scheduling algorithm (e.g. bin packing or work stealing). 

The choice of scheduling algorithms could influence the speedup because inappropriate scheduling algorithm may lead to load balancing problems (see §4.4 for details). Round robin scheduling algorithm works well for DNNs with many blocks which have a similar training time. However, for DNNs like VGG16 where the training time of each layer is vastly different and the total number of replacement blocks is relatively small, naive scheduling such as round robin will likely cause load balancing issues. In this case, more advanced scheduling algorithm such as bin packing or work stealing is needed. The bin packing algorithm considers tasks with varied execution times as items with varied weights and the number of GPUs are viewed as the number of bins. The goal of bin packing scheduling is to allocate tasks to available GPUs in a way that each GPU has balanced workload. More formally, given $n$ bins of unlimited capacity and $m$ tasks of weight $w_{m}$, the bin packing algorithm distributes the $m$ tasks so that for every bin $b_n$ their sum $\sum\limits_{{m\in b}}{w_{m}}$ are as even as possible. In our algorithm, we leverage the bin packing implementation in the python library\cite{binpacking}, which uses the heuristic Worst Fit Decreasing (WFD) approach that first sorts the items to be placed in decreasing order, then places them one by one into the next most empty bin until all tasks are distributed. Bin packing scheduling is very effective but it requires prior knowledge of each task's execution time. Generally speaking, if a model is compressed frequently, it is worthwhile to get such prior knowledge because obtaining the training time once could benefit all compression processes thereafter. However, when prior knowledge is not available or the cost of obtaining such information is too high, bin packing scheduling becomes less ideal. The work stealing scheduling policy can address this issue as the load balancing can be achieved dynamically while running the model. In a work stealing scheduler, each GPU has a queue of tasks. When a GPU runs out of work, it checks the queues of other GPUs and steals their tasks from the tail of the queue. As a result, work stealing achieves dynamic load balancing by distributing the tasks evenly over all GPUs. Our algorithm uses the work stealing scheduling policy implemented by Dask \cite{workstealing}, which is deployed on the Chameleon system \cite{chameleon}. 


Each layer is trained using local loss as described in Section \ref{indep_replacement}, which helps greatly to reduce the amount of subnetwork graph that must be loaded in memory to perform the block distillation. This allows us to use larger batch sizes and perform the distillation on a single GPU. It also results in very fast training time per block because of the limited number of layers that require gradient information during the back prorogation phase. However, this also requires a careful balancing act to load and remove the required components of the computation graph at each part of the training phase. We load our teacher network and save a temporary copy of the teacher subnetwork up until the considered $kth$ block for replacement. Then the GPU memory must be explicitly flushed. After that, the teacher subnetwork and the student block are loaded again and trained for two epochs. Then our algorithm checks how the distillation performs by evaluating the trained accuracy of each replacement block in the student model. Once all GPUs have finished their workload, the main MPI process with rank = 0 does a single gather from all GPUs and layers are compared with a user-defined threshold for accuracy. Only student blocks that exceed the accuracy threshold will be used in the model reassembly process. When the compressed model is assembled, it is fine-tuned for a few epochs to recover some accuracy and returned as the final student model output. 


The performance of a parallel algorithm is usually affected by dependencies, synchronization overhead, load unbalancing, and speed gap between CPUs and GPUs. Our algorithm addresses each bottleneck as follows. First, by extracting local information, our algorithm makes each replacement block training an independent task, which eliminates most dependencies and only requires two synchronizations (one at the beginning to dispatch tasks to GPUs and one at the end to assemble the student model). Second, by leveraging different scheduling algorithms (round robin or bin packing), the workload of each GPU is well balanced. Third, we can adjust the number of threads used for CPU data preprocessing to match the GPU speed on consuming training data. In our experiments, we observed that the recommended default settings in TensorFlow are not ideal so we adjusted the number of CPU threads to achieve better speedup (see §4.3 for details). Successfully tackling all these bottlenecks makes our proposed algorithm a very appealing solution to parallel knowledge distillation. Even better, our algorithm scales automatically on both single machine with multiple GPUs and a multi-node GPU system and this scalability is transparent to users because they do not need to modify their hyper-parameters.         


\section{Experimental Results}

\subsection{System Configuration and Implementation Details}

\subsubsection{Hardware Configuration}
Experiments were conducted on two different systems. The first system is a single server that contains an AMD Ryzen Threadripper 2950x processor (16 physical cores with hyperthreading support for 32 threads), 4 Nvidia RTX 2080TI GPUs, and 128GB of DDR4 Memory in quad channel configuration. The second one contains multiple nodes from the NSF Chameleon system \cite{chameleon}. Each Chameleon node has 192GB of DDR4 Memory, 2 Intel Xeon Gold 6126 Processor (12 physical cores each with hyperthreading support for 24 threads), and a Nvidia RTX6000 GPU. These Chameleon nodes are connected as a cluster using 10Gb network.

\subsubsection{System Profiling Tool}
A system profiling tool is developed to collect real-time information about CPU utilization, CPU power consumption as well as multiple GPUs' utilization and power consumption. To ensure the profiling tool is lightweight, we choose a sampling frequency of 1Hz, which has a negligible impact on the knowledge distillation process. The profiling for GPUs is via the Nvidia Management Library(NVML), which is already well known thus the details are skipped here. We provide more details about profiling the AMD Ryzen Threadripper processor as follows.  

\textbf{Power consumption.} The power consumption data is collected from the Model Specific Register(MSR) files. We can access different registers by seeking for the MSR number as an offset. For example, we can go to each $cpunum$ MSR directory and read 8 bytes starting from the offset $0xC001029A$ to get per CPU core energy. The package energy (i.e. total energy consumed on a single cpu chip) can be obtained from the offset $0xC001029B$. Since the recorded energy value is accumulative, we measure delta energy consumption over a sampling time period to calculate the average power consumption using the following equation: 

        \begin{equation}
            CPU_{avgpower} = \Delta Energy * frequency 
        \end{equation} 


\textbf{CPU utilization.} CPU usage information is obtained from the $/proc/stat/$ file, which holds various information about the kernel activities. To measure CPU utilization, we can sum up $CPU time$ spends on user mode and kernel mode divided by the total $CPU$ time spends within a sampling interval. The average CPU usage can be calculated with the equation below:
        \begin{equation}
            CPU_{avgusage} = \frac {(\Delta CPU_{user} + \Delta CPU_{kernel})} {\Delta CPU_{total}} 
        \end{equation}

\subsubsection{DNN Models}
The VGG16 \cite{simonyan2014very} and ResNet50 \cite{he2016deep} are used to verify the effectiveness of our proposed parallel blockwise distillation algorithm. VGG16 has fewer layers but more parameters in each layer whereas ResNet50 has many more layers but less work is done in a single layer. For both models, we utilize the Keras applications package for implementation. When compressing ImageNet we use the pre-trained weights. For CIFAR10 we train the teacher model from scratch. The other change made to the CIFAR10 implementation of the VGG16 teacher model is we use global average pooling to omit the fully connected layers so as not to overfit the small dataset.

\subsubsection{Datasets}
Two datasets are selected for our experiments. CIFAR10 \cite{krizhevsky2009learning} contains 60,000 RGB images with the size of 32 x 32 pixels. These images are separated in 10 classes and each class has 5,000 images for training and 1,000 for testing. CIFAR10 has been widely used in the model compression field to quickly verify research ideas or build prototypes. ImageNet \cite{imagenet_cvpr09} contains 1.28 million training images and 50,000 validation images in 1,000 classes. This dataset is widely used on image recognition and object detection tasks to verify model performance on a large scale. 

\subsubsection{Data Preprocessing}
We utilize the tf.data library and TensorFlow Datasets for data processing and loading. Standard data augmentation (e.g. rotate, flip, and random crop) is applied to images during training. For models in the Keras applications package, we use their provided preprocess\_input() functions to normalize the input images.

\subsubsection{Training Details}

We train the CIFAR10 implementations for 30 epochs per layer (although most converged within 5-10 epochs) and fine-tune the assembled model for 20 epochs. ImageNet models use provided Keras applications and pre-trained weights. The layers are each trained for roughly the equivalent number of total steps as the CIFAR10 models, which results in about 1.5 epochs on the larger dataset. Models are then fine-tuned for 2 epochs over the full dataset when assembled. ResNet includes many 1x1 convolutional layers which are not replaced in our algorithm. During the fine-tuning stage of ResNet, we freeze all 1x1 convolution layers and fully connected layers so that only the replacement blocks are fine-tuned. Further implementation details can be found on our Github at \url{https://github.com/codestar12/Parallel-Independent-Blockwise-Distillation.}

\subsection{Speedup and Accuracy}

In this section, we demonstrate the experimental results to verify our proposed algorithm can achieve near linear speedup without compromising accuracy.  

Table \ref{tab:cifar-speedup} and \ref{tab:cifar-speedup-multi} depict the speedup and efficiency (the ratio of speedup over the number of GPUs) for VGG and ResNet when using our parallel knowledge distillation algorithm on the CIFAR10 dataset. We observe a 1.92, 2.50, and 3.53 speedup for ResNet and 1.77, 2.45, and 3.08 speedup for VGG when using two, three, and four GPUs respectively on the single AMD server with 4 GPUs. Meanwhile, we achieve even higher speedup (3.87 for 4 GPUs) and efficiency (0.97 for 4 GPUs) for ResNet running on the distributed cluster with work stealing scheduling enabled (see Table 4). Table \ref{tab:imagenet-speedup-multi} shows that our algorithm can also achieve near linear speedup (1.998, 2.92, and 3.64 for two, three, and four GPUs) when training the student VGG using the ImageNet dataset, which is better than the results from the single AMD server (see Table \ref{tab:imagenet-speedup}). This is largely due to the 16 core CPU's inability in the single server to simultaneously process larger volume of ImageNet images for multiple GPUs.  

\begin{table}[h]
\centering
\caption{Speedup and Efficiency of ResNet and VGG on CIFAR10 (Single AMD Server - Bin Packing Scheduling)}
\begin{tabular}{@{}lllll@{}}
\toprule
\# of GPUs                                    & Time (s)                           & Speedup  & Effeciency & DNN Model \\ \midrule
\multicolumn{1}{l|}{4}          & \multicolumn{1}{l|}{2749.81} & 3.53 & 0.96 & ResNet \\
\multicolumn{1}{l|}{3}        & \multicolumn{1}{l|}{3876.03} & 2.50 & 0.83 & ResNet \\
\multicolumn{1}{l|}{2}   & \multicolumn{1}{l|}{5060.97} & 1.92 & 0.88 & ResNet \\
\multicolumn{1}{l|}{1}   & \multicolumn{1}{l|}{9693.24} & 1 & 1 & ResNet \\
\midrule
\multicolumn{1}{l|}{4}        & \multicolumn{1}{l|}{1274.77} & 3.08 & 0.77 & VGG \\
\multicolumn{1}{l|}{3}        & \multicolumn{1}{l|}{1603.12} & 2.45 & 0.82& VGG \\
\multicolumn{1}{l|}{2}        & \multicolumn{1}{l|}{2214.57} & 1.77 & 0.89 & VGG \\ 
\multicolumn{1}{l|}{1}        & \multicolumn{1}{l|}{3920.53} & 1 & 1 & VGG \\\bottomrule
\end{tabular}

\label{tab:cifar-speedup}
\end{table}

\begin{table}[h]
\centering
\caption{Speedup and Efficiency of ResNet and VGG on CIFAR10 (Distributed Cluster - Work Stealing Scheduling)}
\begin{tabular}{@{}lllll@{}}
\toprule
\# of GPUs                                    & Time (s)                           & Speedup  & Effeciency & DNN Model \\ \midrule
\multicolumn{1}{l|}{4}          & \multicolumn{1}{l|}{2094.21} & 3.87 & 0.9685 & ResNet \\
\multicolumn{1}{l|}{2}   & \multicolumn{1}{l|}{4064.43} & 1.996 & 0.998 & ResNet \\
\multicolumn{1}{l|}{1}   & \multicolumn{1}{l|}{8113.087} & 1 & 1 & ResNet \\
\midrule
\multicolumn{1}{l|}{4}        & \multicolumn{1}{l|}{1556.94} & 2.93 & 0.732 & VGG \\
\multicolumn{1}{l|}{2}        & \multicolumn{1}{l|}{2456.59} & 1.856 & 0.927 & VGG \\ 
\multicolumn{1}{l|}{1}        & \multicolumn{1}{l|}{4558.57} & 1 & 1 & VGG \\\bottomrule
\end{tabular}

\label{tab:cifar-speedup-multi}
\end{table}

\begin{table}[h]
\centering
\caption{Speedup and Efficiency of ResNet on ImageNet (Single AMD Server - Bin Packing Scheduling)}
\begin{tabular}{@{}lllll@{}}
\toprule
\# of GPUs                                    & Time (s)                           & Speedup  & Effeciency & DNN Model \\ \midrule
\multicolumn{1}{l|}{4}          & \multicolumn{1}{l|}{13871.72} & 2.75 & 0.69 & ResNet \\
\multicolumn{1}{l|}{3}        & \multicolumn{1}{l|}{16893.19} & 2.26 & 0.75 & ResNet \\
\multicolumn{1}{l|}{2}   & \multicolumn{1}{l|}{20895.00} & 1.83 & 0.91 & ResNet \\ 
\multicolumn{1}{l|}{1}   & \multicolumn{1}{l|}{38171.40} & 1 & 1 & ResNet \\ \bottomrule
\end{tabular}

\label{tab:imagenet-speedup}
\end{table}

\begin{table}[h]
\centering
\caption{Speedup and Efficiency of ResNet on ImageNet (Distributed Cluster - Work Stealing Scheduling)}
\begin{tabular}{@{}lllll@{}}
\toprule
\# of GPUs                                    & Time (s)                           & Speedup  & Effeciency & DNN Model \\ \midrule
\multicolumn{1}{l|}{4}          & \multicolumn{1}{l|}{9234.09} & 3.639 & 0.910 & ResNet \\
\multicolumn{1}{l|}{3}        & \multicolumn{1}{l|}{11517.76} & 2.918  & 0.973 & ResNet \\
\multicolumn{1}{l|}{2}   & \multicolumn{1}{l|}{16821.35} & 1.998 & 0.999 & ResNet \\ 
\multicolumn{1}{l|}{1}   & \multicolumn{1}{l|}{33605.60} & 1 & 1 & ResNet \\ \bottomrule
\end{tabular}

\label{tab:imagenet-speedup-multi}
\end{table}

In terms of accuracy, we compare our algorithm with the progressive blockwise distillation algorithm \cite{wang2018progressive}, which is the state-of-the-art knowledge distillation algorithm. Since the versions of trained VGG and ResNet models used in our experiment are different from those used in \cite{wang2018progressive}, the accuracy of our teacher models is not the same as the accuracy of teacher models used in \cite{wang2018progressive}. For a fair comparison, we highlight the accuracy changes in the last column of Table \ref{tab:accuracy_cifar10}, from which we can see that the accuracy drops by 3.05\% for VGG when using the algorithm proposed in \cite{wang2018progressive} on the CIFAR10 dataset. However, the accuracy degrades by only 0.73\% for VGG and 1.59\% for ResNet when using our algorithm. Table \ref{tab:accuracy_imagenet} shows that the accuracy of our algorithm when using ImageNet dataset decreases by merely 0.6\% for VGG and even increases by 1.91\% for ResNet. 


\begin{table}[h]
\centering
\caption{Comparison of Top-1 Accuracy on CIFAR10 \& CIFAR100}
\begin{tabular}{@{}llll@{}}
\toprule
                                   & Model                           & Top-1   & Accuracy Change \\ \midrule
\multirow{6}{*}{CIFAR10}          & \multicolumn{1}{|l|}{Teacher VGG \cite{wang2018progressive}} & 86.61\% &   \\
                                 & \multicolumn{1}{|l|}{Student VGG \cite{wang2018progressive}} & 83.56\% & -3.05\%  \\
                                 & \multicolumn{1}{|l|}{Teacher VGG Ours} & 87.32\% &  \\
                                                                  & \multicolumn{1}{|l|}{Student VGG Ours} & 86.59\% & -0.73\%  \\
                                 & \multicolumn{1}{|l|}{Teacher ResNet} & 90.36\% &  \\

                                 & \multicolumn{1}{|l|}{Student ResNet} & 88.77\% & -1.59\%  \\
                                 \bottomrule
\multirow{4}{*}{CIFAR100}          & \multicolumn{1}{|l|}{Teacher ResNet \cite{gao2018embarrassingly}} & 71.21\% &   \\
                                 & \multicolumn{1}{|l|}{Student ResNet \cite{gao2018embarrassingly}} & 70.77\% & -0.44\%  \\
                                 & \multicolumn{1}{|l|}{Teacher ResNet Ours} & 72.14\% &  \\
                                                                  & \multicolumn{1}{|l|}{Student ResNet Ours} & 70.53\% & -1.61\%  \\

                                 \bottomrule
\end{tabular}

\label{tab:accuracy_cifar10}
\end{table}

\begin{table}[h]
\centering
\caption{Comparison of Top-1 Accuracy on Imagenet}
\begin{tabular}{@{}llll@{}}
\toprule
                                   & Model                           & Top-1   & Accuracy Change\\ \midrule
\multirow{6}{*}{ImageNet}          & \multicolumn{1}{|l|}{Teacher VGG \cite{wang2018progressive}} & 68.28\% &   \\
                                    & \multicolumn{1}{|l|}{Student VGG \cite{wang2018progressive}} & 70.28\% & +2.00\%  \\
                                 & \multicolumn{1}{|l|}{Teacher VGG Ours} & 65.7\% &  \\
                                 & \multicolumn{1}{|l|}{Student VGG Ours} & 65.1\% & -0.6\%  \\
                                 & \multicolumn{1}{|l|}{Teacher ResNet} & 70.56\% &  \\
                                 & \multicolumn{1}{|l|}{Student ResNet} & 72.47\% & +1.91\%  \\
                                 \bottomrule
\end{tabular}

\label{tab:accuracy_imagenet}
\end{table}

\begin{figure*}[ht!]
     \centering
     
    \begin{subfigure}[b]{0.36\textwidth}
         \centering
         \includegraphics[width=\textwidth]{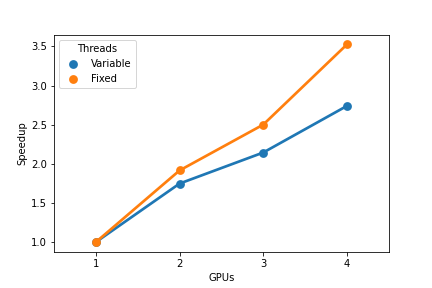}
         \caption{Speedup of ResNet on CIFAR10 }
         \label{fig:Res_speedup}
     \end{subfigure}
     \hfil
    \begin{subfigure}[b]{0.36\textwidth}
         \centering
         \includegraphics[width=\textwidth]{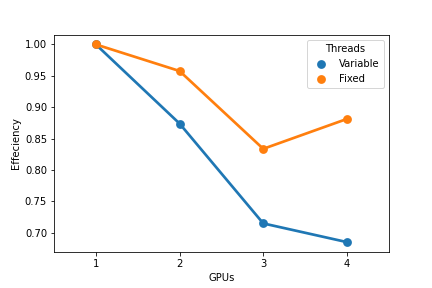}
         \caption{Efficiency of ResNet on CIFAR10 }
         \label{fig:Res_eff}
    \end{subfigure}
     \begin{subfigure}[b]{0.36\textwidth}
         \centering
         \includegraphics[width=\textwidth]{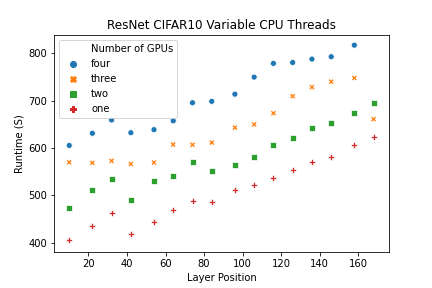}
         \caption{ResNet CIFAR10 layer runtimes with variable number of threads}
         \label{fig:Res_layer_Var}
     \end{subfigure}
     \hfil
     \begin{subfigure}[b]{0.36\textwidth}
         \centering
         \includegraphics[width=\textwidth]{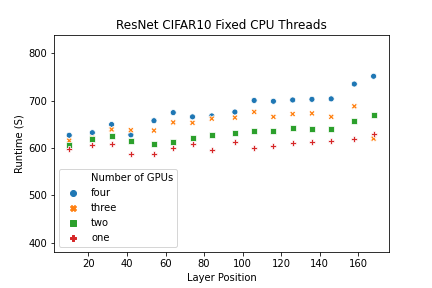}
         \caption{ResNet CIFAR10 layer runtimes with fixed number of threads}
         \label{fig:Res_layer_fixed}
     \end{subfigure}

    \caption{Comparison of ResNet speedup when using fixed or variable number of threads to preprocess data.}\label{fig:Resnet_CIFAR}
\end{figure*}

\subsection{Impact of Data Preprocessing and Hyperthreading}

While training the student model on ResNet using CIFAR10 on the single AMD server, we notice that the speedup and efficiency of our algorithm are below our expectation when using three or four GPUs, as shown in the blue lines of Figure \ref{fig:Res_speedup} and \ref{fig:Res_eff}. This is counter-intuitive because all hyper-parameters and the scheduling algorithm remain unchanged. 

To identify the root cause of this issue, we record timing information for each layer and the entire distillation process. Figures \ref{fig:Res_layer_Var} and \ref{fig:Res_layer_fixed} show the runtime of each individual layer with the X-axis being position in the model and the y-axis being runtime in seconds. We can observe that the time to complete the compression process for a single layer increases progressively when we add additional GPUs. Since the GPU side does not change at all, we start to investigate the preprocessing on the CPU side. It turns out that the number of allowed threads per GPU instance to prepare the image data has a large impact on speedup. More specifically, TensorFlow recommends setting\mintinline{python}{ num_parallel_call} for a \mintinline{python}{tf.data} pipeline to \mintinline{python}{tf.data.experimental.AUTOTUNE}. This aims to achieve high throughput by trying to make the best possible use of available resources. It results in our fastest single GPU total runtime and layer runtime. However, when more threads (especially with hyperthreading enabled) are created to simultaneously preprocess data for multiple GPUs, these threads will fight for shared resources (e.g. cache and the  same  physical  CPU  core)  and  slow  down  the  overall distillation process.

We resolve this issue by setting the number of parallel calls to a fixed number (specifically 4 in our experiments) and adding the option \mintinline{python}{max_intra_op_parallelism = 1}. This indirectly disables hyperthreading in our system (because a maximum number of 16 threads can be created for the 16 available CPU cores) and alleviates the resource contention problem. Figure \ref {fig:Res_layer_fixed} plots the runtime of each layer using fixed number of threads and the orange lines in Figures \ref{fig:Res_speedup} and \ref{fig:Res_eff} demonstrate the improved speedup and efficiency.

\subsection{Impact of Load Balancing}

\begin{figure*}[ht!]
     \centering
     
    \begin{subfigure}[b]{0.36\textwidth}
         \centering
         \includegraphics[width=\textwidth]{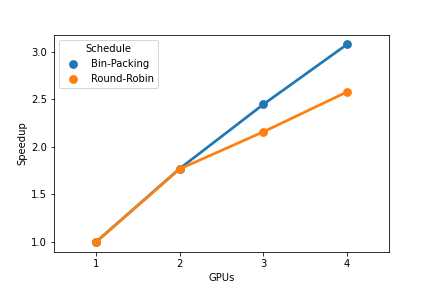}
         \caption{Speedup of VGG16 on CIFAR10}
         \label{fig:vgg_speedup}
     \end{subfigure}
     \hfil
    \begin{subfigure}[b]{0.36\textwidth}
         \centering
         \includegraphics[width=\textwidth]{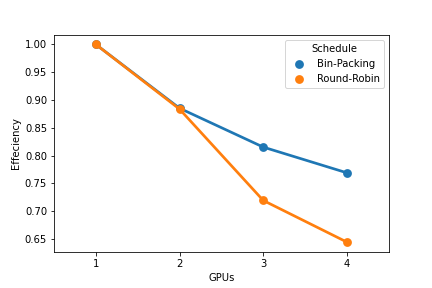}
         \caption{Efficiency of VGG16 on CIFAR10}
         \label{fig:vgg_eff}
    \end{subfigure}
     \begin{subfigure}[b]{0.36\textwidth}
         \centering
         \includegraphics[width=\textwidth]{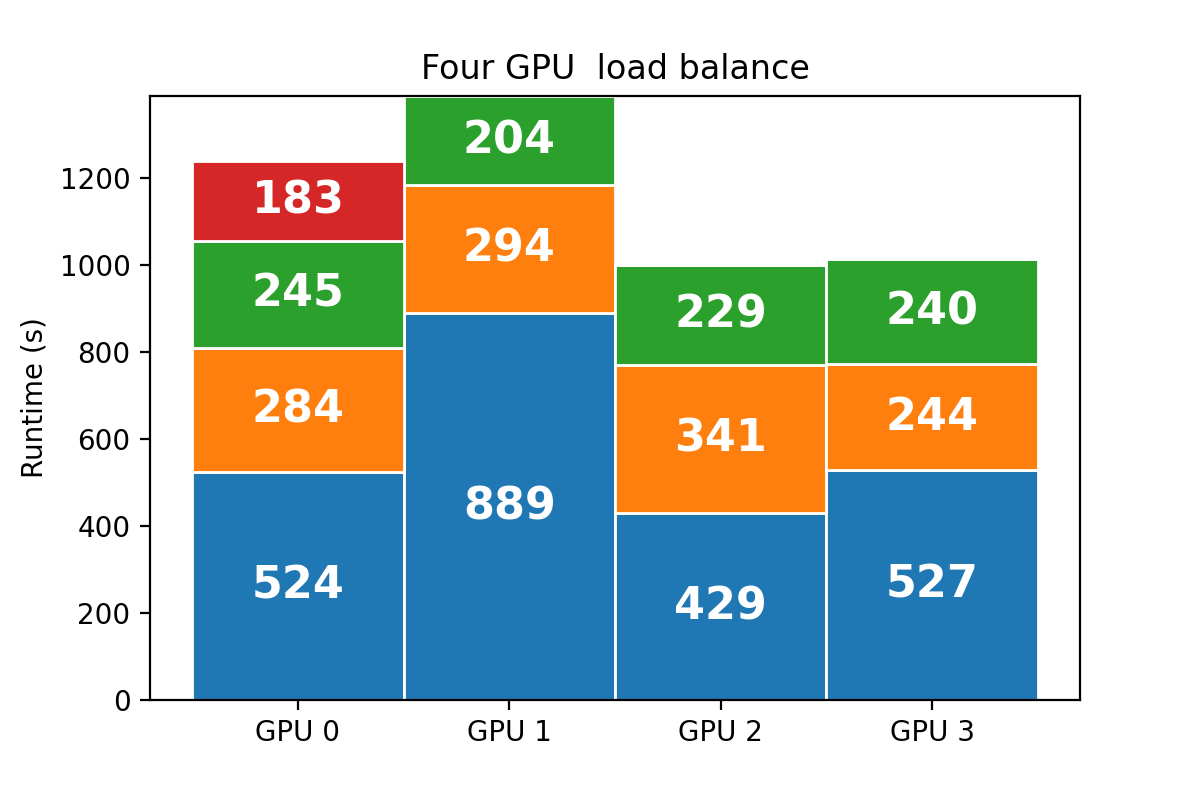}
         \caption{Round Robin task allocation for VGG16}
         \label{fig:four_gp_round_robin}
     \end{subfigure}
     \hfil
     \begin{subfigure}[b]{0.36\textwidth}
         \centering
         \includegraphics[width=\textwidth]{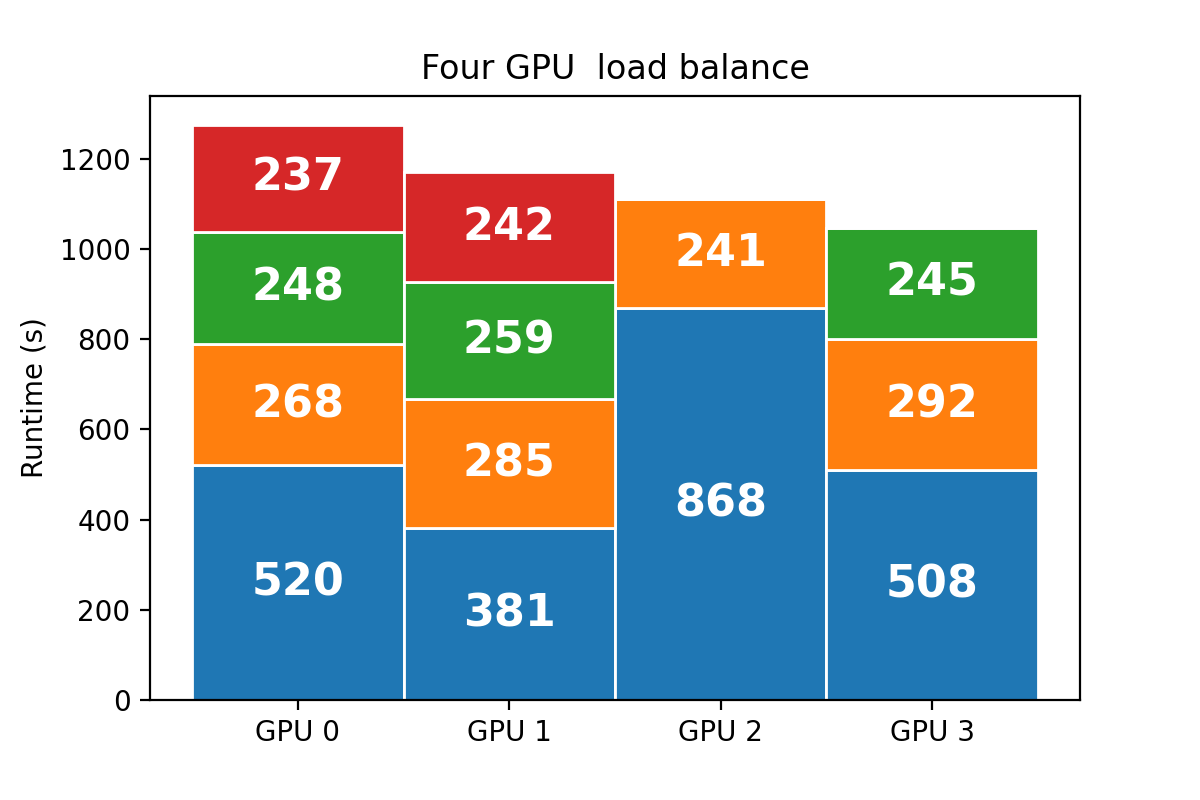}
         \caption{Bin Packing task distribution for VGG16}
         \label{fig:four_gpu_bin}
     \end{subfigure}

    \caption{Comparison of VGG16 speedup when using Round Robin or Bin Packing to schedule layers.}\label{fig:VGG_CIFAR}
\end{figure*}

Another important factor that affects speedup is load balancing, which is largely determined by the characteristics of tasks and how these tasks are scheduled for execution. The naive round robin scheduling achieves good load balancing for ResNet because tasks in ResNet have similar runtimes. However, for DNNs with much fewer blocks to replace and larger variation on feature map size like VGG16, naive scheduling can seriously impact speedup. Figure \ref{fig:four_gp_round_robin} shows how round robin scheduling assigns the 13 VGG tasks (the number inside each task indicates its runtime in seconds) to four GPUs, from which we can clearly see that the workload of GPU0 and GPU1 is much higher than the workload of GPU2 and GPU3. On the other hand, the bin packing scheduling allocates tasks more evenly (see Figure \ref{fig:four_gpu_bin}), which greatly increases the speedup of 4 GPUs from around 2.5 to 3.0 and efficiency from 0.65 to 0.8. However, bin packing scheduling requires priori knowledge about the run-time of each parallel task, which may not be available. In that case, work stealing scheduling can be used to achieve load balancing dynamically.      



\subsection{Energy savings}


Energy consumption is another concern when training and compressing large DNNs. Since DNNs are often trained and evaluated frequently in production applications, even relatively small improvements in energy efficiency can lead to a large reduction in energy and carbon emission over time. In this section, we use the greenup metric to compare the energy consumption of our parallel compression algorithm with the serial implementation. Greenup was first introduced by Abdulsalam et al. in \cite{abdulsalam2015using} and defined as follows:  

\begin{equation}
    Greenup = Energy_{serial} / Energy_{parallel}
\end{equation}

Since the total energy consumption is the accumulated product of runtime over power, all factors that can affect runtime or power can influence total energy consumption. For example, the Dynamic Voltage and Frequency Scaling (DVFS) techniques have been enabled on both CPU and GPU as default settings, which constantly try to reduce power when workload drops. Other factors may have conflicting effects on runtime and power. For example, using more GPUs to achieve larger speedup will help reduce runtime but will also increase both the CPU power and GPU power.     

Table \ref{tab:cifar-greenup} summarizes the total energy and greenup when running our parallel algorithm on the single AMD server using different numbers of GPUs on ResNet and VGG respectively. Our algorithm achieves a maximum of 1.29x greenup (i.e. 29\% energy savings) on ResNet and a maximum of 1.19x greenup (i.e. 19\% energy savings) on VGG, both when using four GPUs. The energy savings mostly come from dramatically shortened training time, like we have discussed in Table \ref{tab:cifar-speedup}.

\begin{table}[h]
\centering
\caption{Greenup of ResNet on CIFAR10 }
\begin{tabular}{@{}lllll@{}}
\toprule
\# of GPUs                                                           & Energy  (kJ) & Greenup & DNN Model\\ \midrule
\multicolumn{1}{l|}{4}           & 977.39 & 1.29 & ResNet \\
\multicolumn{1}{l|}{3}         & 1067.82 & 1.18 & ResNet \\
\multicolumn{1}{l|}{2}    & 1106.68 &  1.18 & ResNet \\ 
\multicolumn{1}{l|}{1}    & 1263.43 &  1 & ResNet \\ 
\midrule
\multicolumn{1}{l|}{4}         & 893.44 & 1.19 & VGG \\
\multicolumn{1}{l|}{3}         & 929.30 & 1.14 & VGG \\
\multicolumn{1}{l|}{2}         & 950.70 & 1.12 & VGG \\ 
\multicolumn{1}{l|}{1}         & 1061.98 & 1 & VGG \\
\bottomrule
\end{tabular}

\label{tab:cifar-greenup}
\end{table}

We also notice that VGG has less energy reduction than ResNet. It is partially because VGG16 has fewer tasks that can be evenly distributed on multiple GPUs, as illustrated in Figures \ref{fig:VGG_CIFAR} (c) and (d). As a result, GPUs that finish their tasks much earlier would have to stay idle waiting for other GPUs to complete. In our experiments, we find that idle devices also consume a considerable amount of energy over time without contributing useful work. Moreover, since the CIFAR10 dataset is rather small for the distillation process, the overall system utilization is not high ($<$ 60\% most of the time). For distillation on larger datasets (e.g. ImageNet), the execution time of each task will be much longer and system utilization will be higher. Consequently, more energy savings are expected because the load imbalancing issue and the DVFS techniques will generate less impact on greenup.  



\section{Conclusions and Future Work}

Knowledge distillation is a promising technique to compress large deep neural networks (DNNs) by replacing the complex sub-networks in the teacher model by simplified sub-networks in the student model. However, existing knowledge distillation algorithms take long time to train. In this paper, we propose a novel parallel blockwise distillation algorithm, which can significantly reduce the training time and energy consumption of the distillation process. The experimental results running on an AMD server with four Geforce RTX 2080Ti GPUs show that our algorithm can achieve 3x speedup plus 19\% energy savings on VGG distillation, and 3.5x speedup plus 29\% energy savings on ResNet distillation, both with negligible accuracy loss. The speedup of ResNet distillation can be further improved to 3.87 when using four RTX6000 GPUs in a distributed cluster. In addition, our method can leverage different scheduling algorithms (e.g. bin packing or work stealing) based on the nature of the target DNNs to achieve good load balancing. More importantly, our algorithm can scale automatically and transparently when more GPUs are available without requiring users to tune their hyper-parameters. 

Our current work can be further extended in two directions. First, we confirm that our method works well on the convolutional layers of VGG and ResNet in this work. We believe it can be applied more broadly and future work can be done to evaluate its effectiveness on other models or different types of layers.  Second, since our method uses only local loss and does not require labeled data, we plan to further investigate if our method can work with unlabeled data and unsupervised learning.

\section{Acknowledgement}

We would like to thank the anonymous reviewers for their valuable comments. The work reported in this paper is supported by the NSF Grant No. CNS-1908658, the NSF sponsored Chameleon system \cite{chameleon}, and the Texas State University Research Enhancement Program. The opinions expressed in this paper are from the authors and do not reflect the views of the sponsors.


%

\ifCLASSOPTIONcaptionsoff
  \newpage
\fi



%
\bibliographystyle{unsrt}
\bibliography{bare_jrnl_compsoc}

%



\begin{IEEEbiography}[{\includegraphics[width=1in,height=1.25in,clip,keepaspectratio]{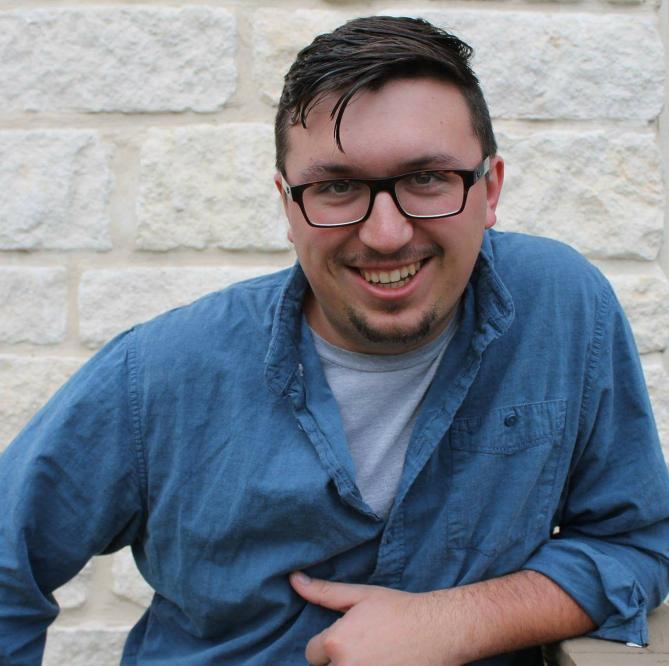}}]{Cody Blakeney}recieved his B.S. in Computer Science from Texas State University in 2018. He is now a Ph.D Student at Texas State University. His research interests are Mobile/Edge Computer Vision and AI.
\end{IEEEbiography}

\begin{IEEEbiography}[{\includegraphics[width=1in,height=1.25in,clip,keepaspectratio]{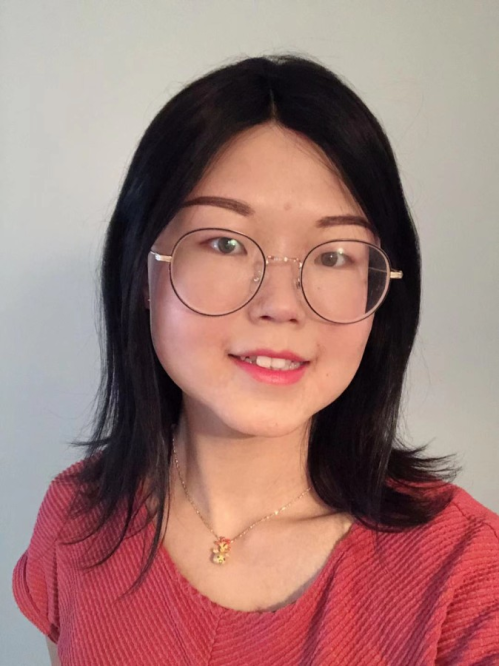}}]{Xiaomin Li} received her B.S. in Computer Science from Lanzhou University, China in 2018. She is now a Ph.D. student at Texas State University. Her research interests are mainly on green AI development, energy-efficient deep neural networks design, and machine learning implementation on recourse-constraint systems. Other than that, she also has strong backgrounds in high-performance computing, parallel programming, and GPU programming. 
\end{IEEEbiography}

\begin{IEEEbiography}[{\includegraphics[width=1in,height=1.25in,clip,keepaspectratio]{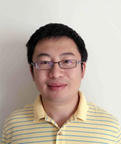}}]{Yan Yan} received his Ph.D. in computer science from the University of Trento and is an Assistant Professor of the Computer Science Department at Texas State University. His research interests include computer vision, machine learning, and multimedia. He has been PC members for several major conferences and reviewers for referred journals in computer vision and multimedia.
\end{IEEEbiography}

\begin{IEEEbiography}[{\includegraphics[width=1in,height=1.25in,clip,keepaspectratio]{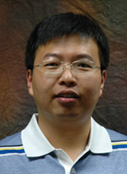}}]{Ziliang Zong} received his Ph.D. degree from Auburn University and is an Associate Professor of the Computer Science Department at Texas State University. His research interests include energy-efficient computing and systems, big data analytics, machine learning, and edge computing. He serves as the Associate Editor of the Sustainable Computing Journal, co-chairs and committee members of numerous conferences and workshops, and reviewers for referred journals and conferences in high performance computing, green computing, and edge computing.
\end{IEEEbiography}




\end{document}